\documentclass[letterpaper, 10 pt, conference]{ieeeconf}  % Comment this line out
\IEEEoverridecommandlockouts                              % This command is only
\usepackage{graphicx}
\usepackage{amsmath}
\usepackage{cite}
\usepackage{url}

\usepackage{times}
\usepackage{fancyhdr,graphicx,amsmath,amssymb}
\usepackage[ruled,vlined]{algorithm2e}

\usepackage{algpseudocode}
\usepackage{soul}

\usepackage[font=footnotesize,belowskip=-0.5pt,aboveskip=0pt]{caption}
\usepackage{bigstrut}
\usepackage[utf8]{inputenc}
\usepackage{array}
\usepackage{multirow}
\usepackage{tabularx}
\usepackage{bigstrut}
\usepackage{makecell}
\usepackage{tabu}
\usepackage{bm}
\usepackage{caption}

\usepackage{amsmath}          
\usepackage{graphics}
\usepackage{graphicx}
\usepackage{float}
\usepackage{caption}
\usepackage{subcaption}
\usepackage{dblfloatfix}
\usepackage{calc}
\usepackage{url}
\usepackage{textcomp}
\usepackage{balance}
\usepackage{siunitx}
\usepackage{mdwlist}
\usepackage{comment}

\usepackage{gensymb}

\usepackage{array}
\usepackage{multirow}
\usepackage{tabularx}
\usepackage{booktabs}

\usepackage{bigstrut}
\usepackage{fixltx2e}

\usepackage{color}
\usepackage{hhline}
\usepackage[dvipsnames]{xcolor}

\usepackage[capitalize]{cleveref}

\begin{document}

\title{\LARGE
Initial Analysis of Data-Driven Haptic Search for the Smart Suction Cup
}

\author{Jungpyo Lee$^{1}$, Sebastian D. Lee$^{1}$, Tae Myung Huh$^{2}$, Hannah S. Stuart$^{1}$% <-this % stops a space
\thanks{$^{1}$ Embodied Dexterity Group, Dept. of Mechanical Engineering, University of California Berkeley, Berkeley, CA, USA.}
\thanks{$^{2}$ The Dept. of Computer and Electrical Engineering, University of California Santa Cruz, Santa Cruz, CA, USA.}%
\thanks{Coorespondence to H. Stuart: hstuart@berkeley.edu.}
}

% The paper headers
\maketitle

%%%%%%%%%%%%%%%%%%%%%%%%%%%%%%%%%%%%%%%%%%%%%%%%%%%%%%%%%%%%%%%%%%%%%%%%%%%%%%%%

% You can make your own Tex to make your paragraph or sections.
\section{Introduction}

Suction cups offer a useful gripping solution, particularly in industrial robotics and warehouse applications%, where they can reduce the reliance on human labor for pick-and-place tasks, thus improving productivity and enabling continuous operations
\cite{hashimoto2017current, perez2016robot}. Vision-based grasp algorithms, like Dex-Net\cite{mahler2016dex}, show promise but struggle to accurately perceive dark or reflective objects, sub-resolution features, and occlusions\cite{mahler2018dex}, resulting in suction cup grip failures. %These issues highlight t
%These represent ongoing challenges in grasp planning with suction cups, despite advancements in depth sensing and visual planners. 
%Researchers have sought responsive pose adjustment techniques for suction cups, like deployable suction grippers\cite{yoo2023compliant} and actively adaptive suction grasping\cite{liu2020adaptive}. These solutions either employ passive compliance or include transducers within, or in close proximity to, the cup.
In our prior work, we designed the Smart Suction Cup, which estimates the flow state within the cup and provides a mechanically resilient end-effector that can inform arm feedback control through a sense of touch. The design was first introduced in%multi-chamber smart suction cup with pressure-based tactile sensing was first introduced in
\cite{huh2021}. %, showing it is capable of discerning surface textures and aligning with an object's normal. 
We then demonstrated how this cup's signals enable haptically-driven search behaviors for better grasping points on adversarial objects\cite{lee2023haptic}. 

This prior work uses a model-based approach to predict the desired motion direction, which opens up the question: \textit{does a data-driven approach perform better?} This report provides an initial analysis harnessing the data collected in \cite{lee2023haptic}. Specifically, we compare the model-based method with a preliminary data-driven approach to accurately estimate lateral pose adjustment direction for improved grasp success.

\section{Methods}

\subsection{Experimental setup and data acquisition}

\begin{figure}[tbp!]
\centering
	% \vspace{-10pt}
	%\centering
	\includegraphics[width=1\linewidth]{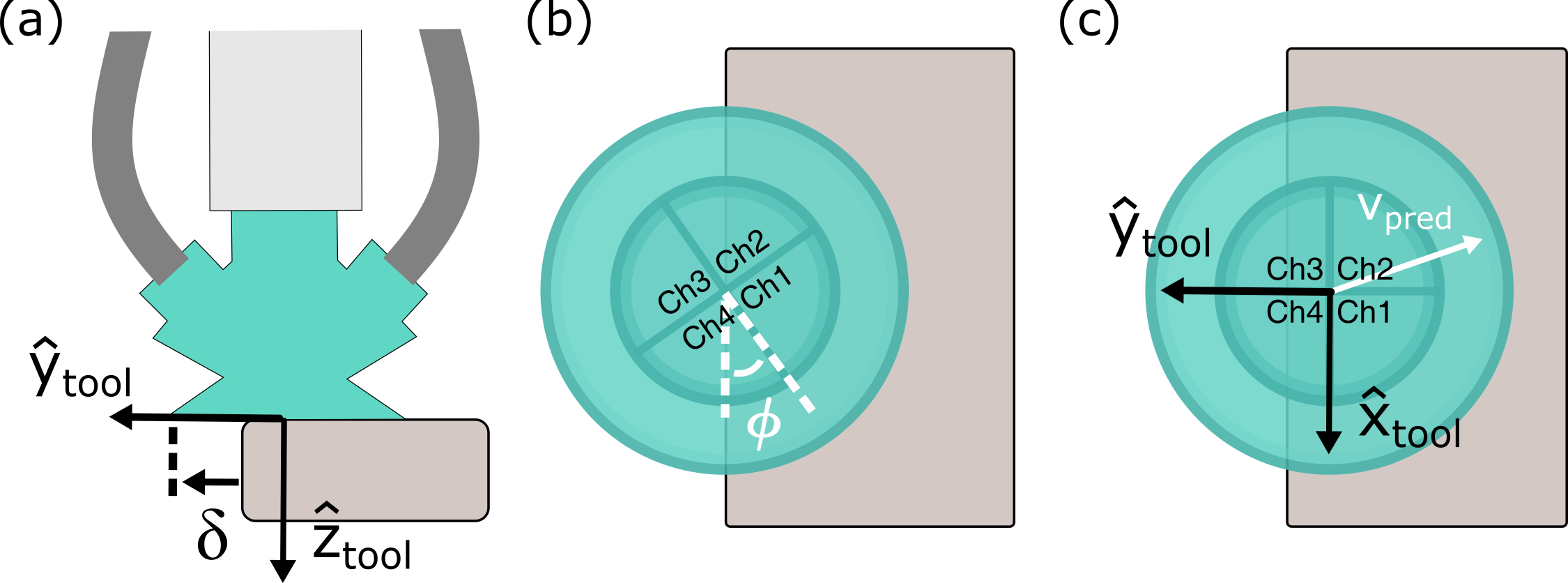}
        \vspace{5pt}
	\caption{The schematic drawing of the experimental setup. (a) A lateral offset and the tool frame ($\delta$). (b) Yaw angle ($\phi$) of the suction cup. (c) A lateral direction vector, $v_{pred}$ is predicted to guide motion. Adapted from \cite{lee2023haptic}.}
	\label{fig:method}
	% \vspace{-15pt}
\end{figure}

We collect a sensor dataset when the suction cup is located near the edge of a plate. The goal is to predict the correct yaw orientation of the cup relative to the edge, $\phi$, as defined in Fig. \ref{fig:method}, in order to translate in the correct direction -- perpendicular to the edge -- to improve gripping. The full experimental procedure, including how lateral offset $\delta$ is changed, is described in \cite{lee2023haptic}. 
%our setup involves positioning and orienting the suction cup concerning the edge of a flat plate, as depicted in \cref{fig:method}. In this context, the lateral offset, denoted as $\delta$, represents the length of the exposed lip, and the orientation is described by the yaw angle ($\phi$) ranging from 0$\degree$ to 360$\degree$ to examine potential asymmetry in pressure sensor responses. To ensure a consistent vertical contact between the flat plate and the suction cup throughout the trials, a constant normal force of 1.5 N is applied at a lateral offset of 0 mm, and then the suction cup's height was maintained as constant throughout a trial. 
We strategically use data for lateral offsets from 7 to 14 mm, the region in which the suction cup fails to grasp the flat plate while measuring meaningful pressure differences between suction cup sensors.
%We systematically vary the lateral offset from 7 to 14 mm, the region in which the suction cup fails to grasp the flat plate while measuring meaningful pressure differences between different chambers. %We change the offset with 1 mm increments. Additionally, we sweep the yaw angle from 0$\degree$ to 360$\degree$ at 5$\degree$ intervals.

\subsection{Model-based prediction}
We predict the yaw angle using the method described in \cite{lee2023haptic}. Briefly, the initial step involves defining vacuum pressure $P_i = P_{atm} - P_{ch.i}$, where i = {1, 2, 3, 4} represents the four pressure sensors and $P_{atm}$ represents atmospheric pressure. Then, the predicted desired motion direction vector $v_{pred}$ is calculated as:
\begin{align}
    \nonumber
    v_{pred} &= ((P_1 + P_4) - (P_2 + P_3)) \hat{x}_{tool} \\
    &+ ((P_3 + P_4) - (P_1 + P_2)) \hat{y}_{tool}
    \label{equ:vector}
\end{align}
We calculate the predicted angle $\phi_{pred}$ from this vector, assuming $v_{pred}$ is perpendicular to the plate's edge.

\subsection{Data-driven prediction}

We implement a conventional Multi-Layer Perceptron (MLP) model that takes the four raw pressure signals from each of the four chambers $P_{ch.i}$ as inputs and generates directional predictions in terms of angle $\phi_{pred}$. This MLP model includes three hidden layers with 16, 32, and 16 neurons, respectively, and employs the Rectified Linear Unit (ReLU) as the activation function for each of these layers.  The optimizer is Root Mean Square Propagation (RMSprop). %The training process involving a Mean Squared Error (MSE) loss function. \
For the learning dataset, we incorporate 25,273 data points, allocating 80\% of these data points for training and 20\% for validation purposes.

\section{Results and Discussion}

\begin{figure}[tbp!]
\centering
	% \vspace{-10pt}
	%\centering
	\includegraphics[width=1\linewidth]{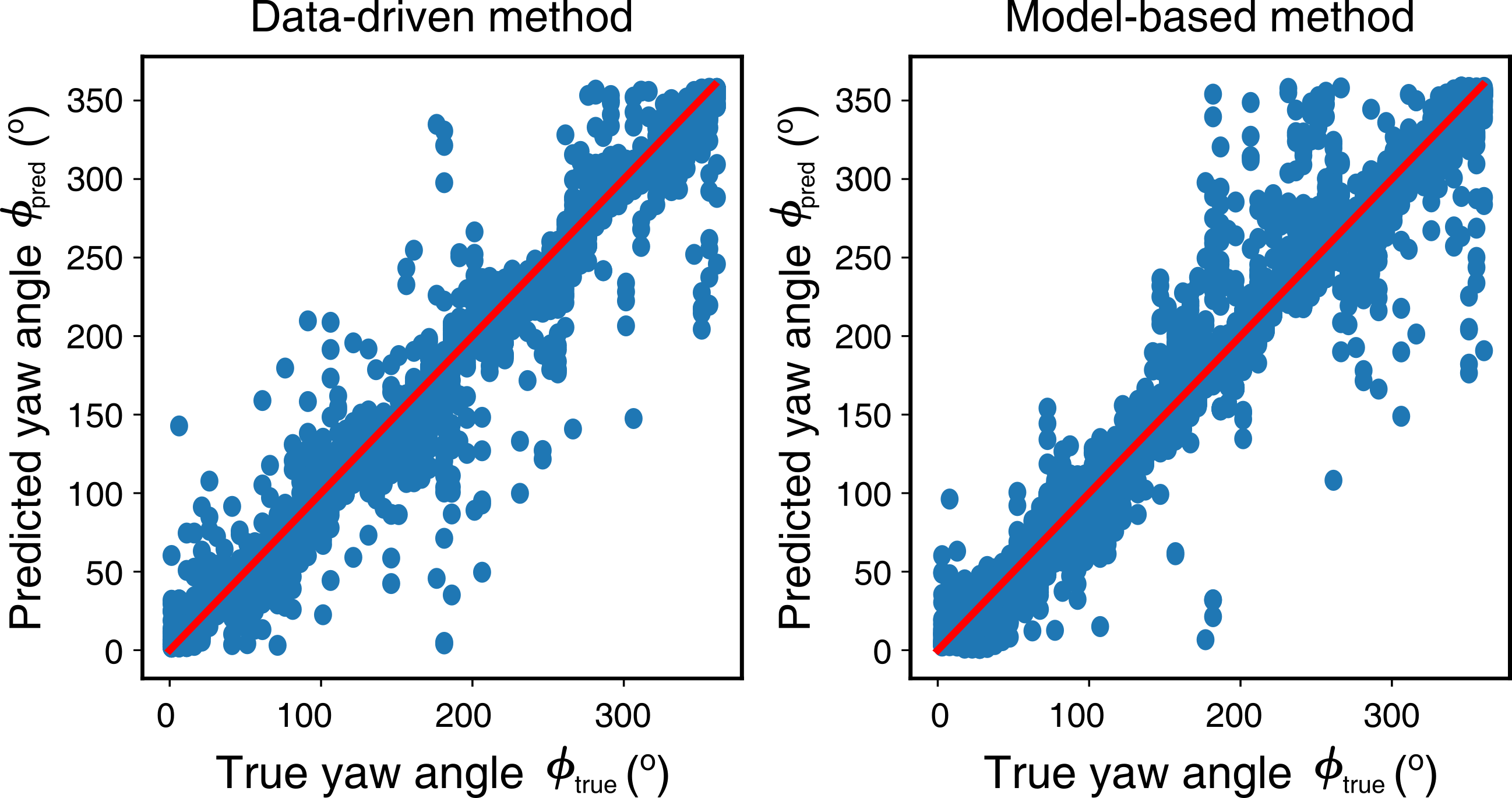}
        \vspace{5pt}
	\caption{The representative results of yaw angle $\phi$ estimation, showing predicted yaw angle and true yaw angle from data-driven (left) and model-based (right) methods.}
	\label{fig:result}
	% \vspace{-15pt}
\end{figure}

The root mean squared error yields a value of 17.23 $\pm$ 0.61$\degree$ from the learning-based prediction, whereas our model-based prediction resulted in 20.49 $\pm$ 0.30$\degree$. Fig. \ref{fig:result} illustrates the predicted angle versus the actual angle for both learning-based and model-based predictions for lateral positioning.
The outcomes indicate that the simple learning-based approach exhibits a slightly enhanced predictive capability for lateral positioning on the smooth, flat plate edge. At the same time, this result also shows our simple model-based method is not far off from the data-driven approach, even across different $\phi$ orientations.

\section{Conclusion}

%Both model-based and data-driven methods hold advantages. 
While the model-based approach tested here requires no training and provides useful results quickly, it does not capture non-idealities in cup manufacturing nor unexpected flow conditions that can lead to errors. Data-driven methods are likely to become important for Smart Suction Cup technology, especially as it is applied to more intricate objects and complex grasping scenarios. However, the efficacy of data-driven methods relies on the availability of suitable datasets -- whether focused on object primitives as in the present work or encompassing unstructured conditions. The integration of physically accurate simulations for Sim2Real translation could streamline this process.

%In this work, our proposed autonomous haptic search with the Smart Suction Cup method enables the suction cup to adjust to a successful pose for suction grasping, effectively increasing tolerance to cartesian positioning or tilt errors induced by errors from a vision-based grasp planner. Future work includes using learning-based approaches for determining the amount of lateral and rotational movements as well as the ratio for mixing these two motion primitives.

\section*{Acknowledgement}
This work is supported by InnoHK of the Government of
the Hong Kong Special Administrative Region via the Hong Kong Centre for Logistics Robotics and by the University of California at Berkeley. The authors thank the members of the Embodied Dexterity Group.

\bibliographystyle{IEEEtran}

\balance  % We need this line for the correct ordering of reference. ( this line is in "balance" Package

\bibliography{IEEEabrv,Biblio}   % Reference are in "Biblio.bib" file.
%\vspace{\baselineskip}

%\input{100_Bio.tex}

\end{document}